\documentclass[letterpaper, 10 pt, conference]{ieeeconf}  

\IEEEoverridecommandlockouts                              
\usepackage[T1]{fontenc}

\usepackage[nolist]{acronym}
\usepackage{comment}
\usepackage{algorithm}
\usepackage{algpseudocode}
\usepackage{url}
\usepackage{cite}
\usepackage{microtype}
\usepackage{todonotes}
\usepackage{booktabs}
\usepackage{multirow}
\usepackage{blindtext}
\usepackage{graphics}
\usepackage{threeparttable}
\usepackage[hang,flushmargin]{footmisc}
\usepackage{flushend} 

\begin{acronym}
\acro{ML}{Machine Learning}
\acro{GPU}{Graphics Processing Units}
\acro{DNN}{Deep Neural Networks}
\acro{CNN}{Convolutional Neural Network}
\acro{BNN}{Bayesian Neural Network}
\acro{TCN}{Temporal Convolutional Network}
\acro{RNN}{Recurrent Neural Networks}
\acro{GNN}{Graph Neural Networks}
\acro{GAT}{Graph Attention Network}
\acro{GCN}{Graph Convolutional Networks}
\acro{GRU}{Gated Recurrent Unit}
\acro{LSTM}{Long Short-Term Memory}
\acro{KF}{Kalman Filters}
\acro{BEV}{Bird's Eye View}
\acro{GAN}{Generative Adversarial Network}
\acro{MTP}{Multi-Modal Trajectory Prediction}
\acro{MLP}{Multilayer Perceptron}
\acro{MAE}{Mean Absolute Error}
\acro{MSE}{Mean Squared Error}
\acro{ADE}{Average Displacement Error}
\acro{FDE}{Final Displacement Error}
\acro{minADE}{Minimum Average Displacement Error}
\acro{minFDE}{Minimum Final Displacement Error}
\acro{VAE}{Variational Autoencoders}
\acro{CVAE}{Conditional Variational Autoencoder}
\acro{AV}{Automated Vehicle}
\acro{MHA}{Multi-Head Attention}
\acro{AD}{Autonomous Driving}
\acro{AV}{Autonomous Vehicle}
\acro{DL}{Deep Learning}
\acro{RL}{Reinforcement Learning}
\acro{MR}{Miss Rate}
\acro{IL}{Imitation Learning}
\acro{MDP}{Markov Decision Process}
\acro{POMDP}{Partially Observable Markov Decision Process}
\acro{RSSM}{Recurrent State Space Models}
\acro{VAE}{Variational Autoencoder}
\acro{UT}{Unscented Transform}
\acro{GMM}{Gaussian Mixture Model}
\acro{NF}{Normalizing Flow}
\acro{KL}{Kullback-Leibler}
\acro{UT}{Unscented Transform}
\acro{FID}{Fréchet Inception Distance}
\acro{NLL}{Negative Log Likelihood}
\acro{ELBO}{Evidence Lower Bound}
\acro{WTA}{Winner-Takes-All}
\acro{EM}{Expectation Maximization}
\acro{NMS}{Non-Maximum Suppression}
\acro{ID}{In-Distribution}
\acro{OOD}{Out-of-Distribution}
\acro{MBRM}{Model-Based Risk Minimization}
\acro{HD}{High Definition}
\acro{OOD}{Out-of-Distribution}
\acro{MC}{Monte-Carlo}
\acro{MCMC}{Markov Chain Monte Carlo}
\acro{VI}{Variational Inference}
\acro{RIP}{Robust Imitative Planning}
\end{acronym}

\usepackage{amsmath,amsfonts,bm}

\def\eqref#1{equation~\ref{#1}}

\def\1{\bm{1}}

\def\rmH{{\mathbf{H}}}
\def\rmI{{\mathbf{I}}}

\DeclareMathAlphabet{\mathsfit}{\encodingdefault}{\sfdefault}{m}{sl}
\SetMathAlphabet{\mathsfit}{bold}{\encodingdefault}{\sfdefault}{bx}{n}

\def\gW{{\mathcal{W}}}

\newcommand{\E}{\mathbb{E}}

\newcommand{\R}{\mathbb{R}}

\makeatletter
\let\NAT@parse\undefined
\makeatother

\usepackage[pdfencoding=auto, colorlinks=true]{hyperref} 

\title{\LARGE \bf
Stochasticity in Motion: An Information-Theoretic\\ Approach to Trajectory Prediction
}

\author{Aron Distelzweig$^{1,2}$, Andreas Look$^{1}$, Eitan Kosman$^1$, Faris Janjo\v{s}$^1$, Jörg Wagner$^1$, Abhinav Valada$^2$
\thanks{$^{1}$Bosch Center for Artificial Intelligence, Germany, Israel.}%
\thanks{$^{2}$Department of Computer Science, University of Freiburg, Germany.}%
}

\begin{document}

\maketitle
\thispagestyle{empty}
\pagestyle{empty}

\begin{abstract}
In autonomous driving, accurate motion prediction is crucial for safe and efficient motion planning. To ensure safety, planners require reliable uncertainty estimates of the predicted behavior of surrounding agents, yet this aspect has received limited attention. In particular, decomposing uncertainty into its aleatoric and epistemic components is essential for distinguishing between inherent environmental randomness and model uncertainty, thereby enabling more robust and informed decision-making.
This paper addresses the challenge of uncertainty modeling in trajectory prediction with a holistic approach that emphasizes uncertainty quantification, decomposition, and the impact of model composition.
Our method, grounded in information theory, provides a theoretically principled way to measure uncertainty and decompose it into aleatoric and epistemic components. 
Unlike prior work, our approach is compatible with state-of-the-art motion predictors, allowing for broader applicability.
We demonstrate its utility by conducting extensive experiments on the nuScenes dataset, which shows how different architectures and configurations influence uncertainty quantification and model robustness.

\end{abstract}

\section{Introduction}
\label{sec:introduction}
In a machine learning driven \ac{AD} stack, motion prediction connects environment perception with ego motion planning~\cite{hu2023uniad}. 
The role of a motion predictor is to infer the future motion of relevant traffic agents to the ego agent, ensuring safe and efficient progress toward a goal~\cite{hagedorn2024integration}.
To achieve this, a predictor must tackle several challenges, including imperfect perception, complex interactions between agents, and the multitude of potential actions that each agent could undertake. Addressing these challenges requires a probabilistic approach that incorporates uncertainty into prediction outputs, which is essential to ensure interpretability and build trust in the overall system.

In the \ac{AD} community, the future motion of surrounding traffic agents is often modeled in the form of trajectories.
Thus, probabilistic trajectory prediction involves capturing a distribution $p({y}|{x},\mathcal{D})$ of future trajectories ${y}$ conditioned on contextual data ${x}$ and a dataset $\mathcal{D}$. 
Contextual data ${x}$ usually contains past trajectories of surrounding agents and map information. 
There are different strategies for capturing this multi-modal distribution.
Some methods attempt to directly predict the modes of the distribution along with their associated weights~\cite{gao2020vectornet, kim2021_lapred, deo2022_pgp}. 
Others use a parametric mixture distribution, such as a \ac{GMM}, where the modes correspond to the predicted trajectories~\cite{tolstaya2021identifying, varadarajan2022multipath++, liu2024_laformer, look2023_cheap}.
Alternatively, generative trajectory prediction models use well-known autoencoder or diffusion architectures to model latent variables and draw trajectory samples~\cite{salzmann2020trajectron++, janjovs2023conditional, jiang2023motiondiffuser}.

\begin{figure}
\centering
\includegraphics[width=1.0\linewidth]{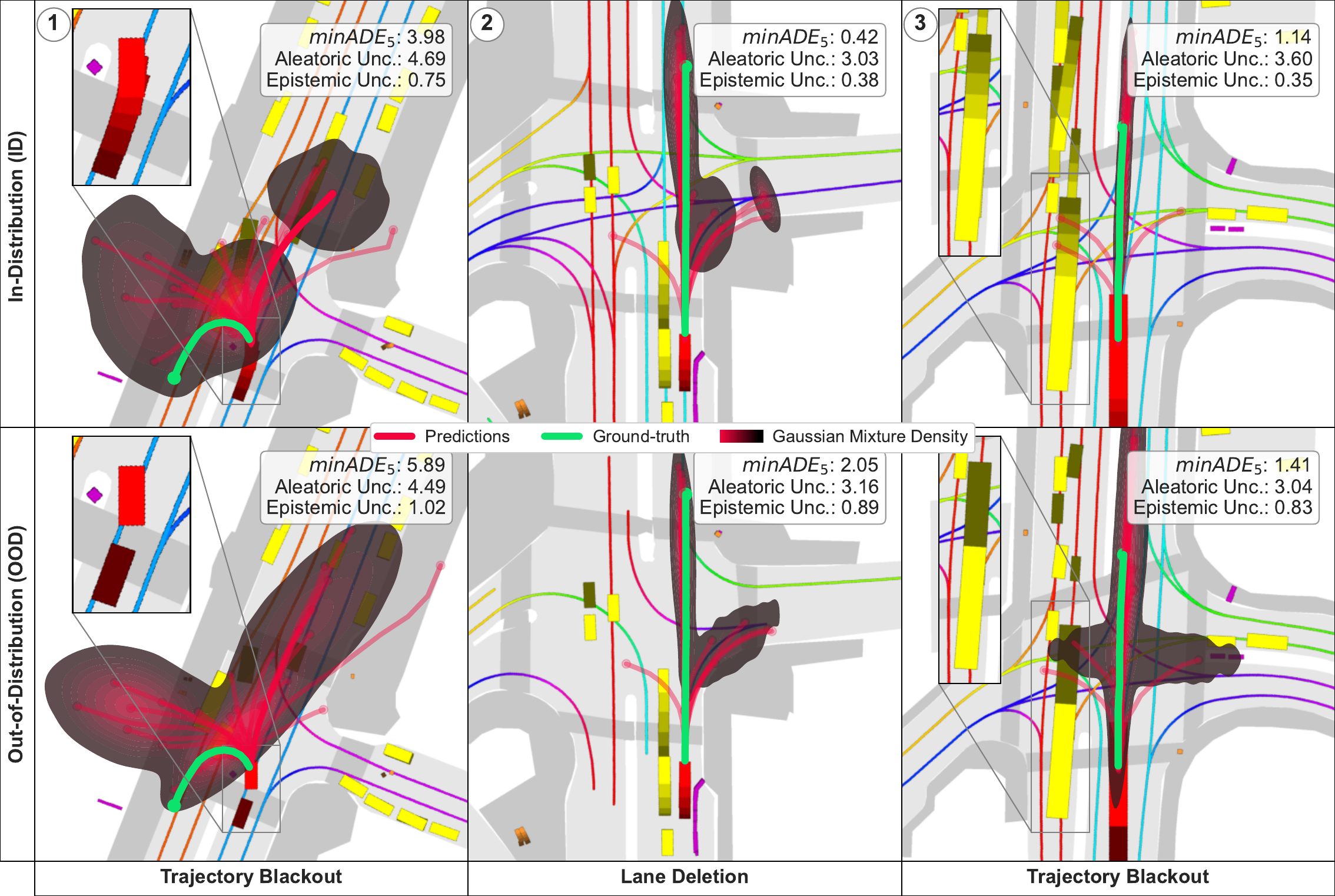}
 \caption{The predictive distribution $p(y|x,\mathcal{D})$ of future trajectories for three example scenarios. The first row shows in-distribution scenarios, while the second row presents \ac{OOD} cases: in \raisebox{.5pt}{\textcircled{\raisebox{-.9pt}{1}}} and \raisebox{.5pt}{\textcircled{\raisebox{-.9pt}{3}}}, segments of the input history have been removed, while in \raisebox{.5pt}{\textcircled{\raisebox{-.9pt}{2}}}, parts of lane information have been removed. Both alterations mimic perception malfunctions. Naturally, prediction error is higher in the second row, indicated by the higher \ac{minADE} metric, see Sec.~\ref{sec:results} for details. 
 Generally, we observe a correlation between \ac{minADE} and total uncertainty. 
 In these examples, epistemic uncertainty serves as a useful indicator for detecting OOD scenarios.}
\label{fig1}
\end{figure}

Most approaches for modeling the distribution of future trajectories in \ac{AD} rely on neural networks.
They are often underspecified by the available data, meaning that no single parameter configuration is favored. 
When considering uncertainty in the model parameters, the predictive distribution $p({y}|{x},\mathcal{D})$ over future trajectories $y$ can be approximated~\cite{kapoor2022on} in the following manner
\begin{align}
\nonumber p(y|x,\mathcal{D}) &= \int p(y|x, \gW)p(\gW|\mathcal{D})d\gW \\
&\approx \int p(y|x, \gW)q(\gW)d\gW,
\label{eq:predictive_dist}
\end{align}
where $\gW$ represents the neural network weights and $p(\gW|\mathcal{D})$ represents the posterior distribution. 
The predictive distribution represents a Bayesian model average, meaning that instead of relying on a single hypothesis with a specific set of parameters, it considers all possible parameter configurations weighted by their posterior $p(\gW|\mathcal{D})$. 
This marginalization process removes the reliance on a single weight configuration in the predictive distribution, resulting in better calibration and accuracy~\cite{wilson2020bayesian}.
Since the exact posterior is intractable, various approximations $q(\gW)$ have been developed, such as variational inference~\cite{graves18_vi}, Dropout~\cite{gal16_dropout}, Laplace approximation~\cite{ritter2018a_laplace}, deep ensembles~\cite{Lakshminarayanan17_ensembles}, or \ac{MCMC} methods~\cite{welling2011_sgld}.

Despite numerous successful approaches to approximating the posterior distribution, the \ac{AD} prediction community has yet to systematically quantify and decompose the uncertainty of trajectory prediction models in a theoretically principled manner~\cite{wilson2020_casebayesiandeeplearning}.
A notable exception is~\cite{itkina2022interpretable}, whose approach is inherently tied to predicting a categorical distribution over fixed trajectories, as done in~\cite{phan2020_covernet} for example. This assumption limits their method, as state-of-the-art predictors do not generate their outputs by ranking a fixed set of trajectories. Overall, the literature gap is surprising given the importance of uncertainty modeling and its ubiquity in other domains~\cite{depeweg2018decomposition,kendall2017uncertainties,abdar2021review}.

In uncertainty modeling, total uncertainty can be categorized into two types: aleatoric and epistemic~\cite{wimmer23a_uq, huellermeier2021uqml}. Aleatoric uncertainty represents inherent variability in the data, such as the equal likelihood of a vehicle turning left or right at a T-junction. This type of uncertainty cannot be reduced, even with additional data.
In contrast, epistemic uncertainty arises from a lack of knowledge and can be reduced by collecting more data~\cite{wimmer23a_uq}. Understanding epistemic uncertainty is valuable in various contexts, such as risk-sensitive reinforcement learning~\cite{depeweg2018decomposition} and out-of-distribution (OOD) detection~\cite{monish23evcenternet,mohan23pods}. By analyzing uncertainty sources, an \ac{AV} can recognize OOD scenarios by detecting increased epistemic uncertainty.
This can serve as a critical signal for a planner that relies on predictions for decision making. Incorporating this information enables planners to make more informed decisions and take precautionary actions in high-uncertainty situations. For example, a vehicle could signal for a human takeover in scenarios with high epistemic uncertainty~\cite{rangesh2022predictingtakeovertimeautonomous}.
Furthermore, uncertainty analysis can be applied not only to the behavior of other agents but also to the \ac{AV}’s own planned trajectory, providing deeper insights into decision-making confidence and potential risks.

In this paper, we address the challenge of modeling the uncertainty of trajectory prediction models within the \ac{AD} domain from a holistic perspective. 
We focus on the quantification and decomposition of uncertainty, as well as the influence of modeling choices related to the approximate posterior $q(\gW)$.
Our method employs an information-theoretic approach~\cite{huellermeier2021uqml}, which quantifies aleatoric uncertainty through conditional entropy and epistemic uncertainty using mutual information.
Fig.~\ref{fig1} shows the predictive distribution $p(y|x,\mathcal{D})$ and its accompanying uncertainty values obtained by our proposed method for different scenarios of the nuScenes dataset~\cite{nuscenes}. 
We summarize our contributions as follows.
\begin{enumerate}
    \item We propose a novel method to quantify and decompose the uncertainty of trajectory prediction models, utilizing conditional entropy and mutual information to measure aleatoric and epistemic uncertainty. 
    \item We analyze the relationship between uncertainty and prediction error in both in-distribution and out-of-distribution scenarios. 
    \item We study how posterior modeling choices impact uncertainty calibration and prediction robustness.
\end{enumerate}


\section{Related Work}
\label{sec:related_work}

Anticipating the future motion of traffic participants is a critical component of autonomous driving systems~\cite{hu2023uniad}. Due to the safety-critical nature of these systems, it's essential to account for uncertainties across the entire prediction stack. For instance, planners need to factor in motion prediction uncertainty to accurately assess the risks associated with various driving maneuvers~\cite{filos2020can}. In the following, we review related work on both motion prediction as well as uncertainty quantification and decomposition.

{\parskip=2pt\noindent\textit{Motion Prediction for Autonomous Driving}:
\label{subsec:related_work_motion_prediction}
The future motion of other traffic participants is influenced by a multitude of observable and unobservable factors, rendering it a challenging modeling task. These factors include, among others, the latent goals and preferences of traffic participants, social norms and traffic rules, complex interactions with surrounding traffic, as well as constraints induced by the static environment~\cite{rudenko2020human}. The shortcomings of a perception system that provides noisy and partial observations pose an additional challenge. These challenges necessitate a probabilistic formulation of the task to adequately model the uncertain and multi-modal nature of future motion. In general, prediction models typically consist of two components: a behavior backbone that encodes the traffic scene and a decoder that models the predictive distribution. We highlight various implementations of the two components below.}

Early prediction approaches~\cite{phan2020_covernet} propose encoding the past trajectory of observed traffic participants and the elements of the static environment (e.g., lane boundaries, crosswalks, traffic signs) by rendering the scene in a semantic bird's eye view image and applying well-established convolutional neural networks. Such image-based representations of the scene have largely been replaced by vectorized representations~\cite{gao2020vectornet,kim2021_lapred,deo2022_pgp,nayakanti2023_wayformer}. In a vectorized representation, all entities of the static and dynamic environment are approximated by a sequence of vectors. Models for sequential data, such as temporal convolutional networks~\cite{oord2016wavenet} or recurrent neural networks are used to encode the sequences and interactions between entities are modeled using pooling operations, graph neural networks, or transformers.

The future motion of traffic participants is typically characterized by a sequence of states over multiple time steps, known as trajectories~\cite{ngiam2021scene, varadarajan2022multipath++}. Several strategies are employed to capture the highly multi-modal distribution over trajectories conditioned on the encoded scene. Many approaches represent the distribution by a set of trajectories with associated mode probabilities. The trajectories are either directly regressed by the model~\cite{cui2019multimodal,liang2020learning,kim2021_lapred,deo2022_pgp} or fixed a beforehand~\cite{phan2020_covernet}. 
In the case of predefined trajectories, the model is responsible for selecting the most likely ones.
Other approaches use parametric mixture distributions, such as \acp{GMM}~\cite{khandelwal2020if,tolstaya2021identifying, varadarajan2022multipath++} or mixtures of Laplacians~\cite{liu2024_laformer}. Alternatively, generative models such as conditional variational autoencoders~\cite{lee2017desire,bhattacharyya2019conditional,salzmann2020trajectron++,janjovs2023conditional}, generative adversarial networks~\cite{gupta2018social,huang2020diversitygan,gomez2022exploring}, diffusion models~\cite{jiang2023motiondiffuser}, or normalizing flows~\cite{scholler2021flomo} model the trajectory distribution via latent variables.

{\parskip=2pt\noindent\textit{Uncertainty Modeling, Decomposition and Quantification}: The majority of current trajectory prediction models solely account for aleatoric uncertainty by modeling a probability distribution over the output space~\cite{varadarajan2022multipath++}. To incorporate epistemic uncertainty in a theoretically sound manner, one can adopt a Bayesian framework~\cite{kendall2017uncertainties, depeweg2018decomposition, wilson2020bayesian, wilson2020_casebayesiandeeplearning}. 
A Bayesian neural network assumes a distribution over the network weights instead of a point estimate to account for the lack of knowledge about the data generation process~\cite{huellermeier2021uqml, jospin2022hands}. Since analytically evaluating the posterior distribution over the weights is intractable for modern neural networks, approximate inference techniques such as \ac{VI} or forms of \ac{MCMC} must be considered~\cite{jospin2022hands}.
Due to its simplicity, \ac{MC} Dropout, which can be interpreted as an approximate \ac{VI} method~\cite{gal16_dropout}, is used by many perception approaches in \ac{AD}~\cite{kendall2017uncertainties,abdar2021review} and is also employed as one of two methods in~\cite{janjovs2023bridging} for modeling solely the epistemic uncertainty of a trajectory predictor. Another well-established approach to account for epistemic uncertainty are deep ensembles~\cite{Lakshminarayanan17_ensembles,jospin2022hands,wilson2020bayesian}. 
Prior work~\cite{filos2020can} uses deep ensembles to approximate the posterior distribution in their epistemic uncertainty-aware planning method. 
We apply MC Dropout as well as deep ensembles to approximate the uncertainty over network weights and systematically assess their performance in the context of trajectory prediction.}

A common information-theoretical measure for the uncertainty is the entropy of the predictive distribution as a measure of the total uncertainty, which can be additively decomposed into the conditional entropy and mutual information, representing a measure of aleatoric and epistemic uncertainty~\cite{depeweg2018decomposition, huellermeier2021uqml, wimmer23a_uq}. Alternative measures based on variance are proposed in~\cite{depeweg2018decomposition}. While variance-based measures are suitable in cases where the predictive distribution is a uni-modal Gaussian, it is less suitable for multi-modal outputs such as trajectories. 
Our approach thus relies on entropy-based measures to quantify the uncertainty of trajectory prediction models. However, variance can be useful in other contexts; \cite{gilles2022uncertainty} uses the variance of the predicted heat map over future positions as an uncertainty measure. Another variance-based uncertainty heuristic is proposed by~\cite{filos2020can} in the related field of motion planning for \ac{AD}. This approach however only quantifies the epistemic uncertainty. Other methods learn proxy measures for the uncertainty of a trajectory prediction model without a proper decomposition: the approach in~\cite{pustynnikov2021estimating} trains separate models while~\cite{janjovs2023bridging} and~\cite{wiederer2023joint} include additional heads with auxiliary tasks. 

To the best of our knowledge, we are the first to offer a thorough and theoretically sound approach for modeling, decomposing, and quantifying uncertainties in trajectory prediction as a solid basis for future downstream applications. Existing approaches in the literature either fail to address all three aspects or rely on heuristics. 


\section{Method}
\label{sec:approach}

This section details our method for decomposing the uncertainty into aleatoric and epistemic parts. 
We start by defining the problem of uncertainty decomposition in trajectory prediction in Sec.~\ref{sec:problem_statement}.
Then, in Sec.~\ref{sec:solution}, we describe our approach for approximating these uncertainties using a \ac{MC} method. 
Finally, we discuss the limitations of our approach with possible avenues to address these in Sec.~\ref{sec:limitations}.

\subsection{Problem Statement}\label{sec:problem_statement}
Our method focuses on uncertainty quantification in trajectory prediction tasks. 
The problem is defined as predicting the future trajectory of a target agent in a driving scene based on current observations. 
Formally, let $x \in \R^{T_{in} \times F_{in}}$ represent the past features of an agent, where $T_{in}$ is the number of observed timesteps and $F_{in}$ denotes the number of input features, such as coordinates, velocities, accelerations, and other relevant data.
In line with recent trajectory prediction literature \cite{deo2022_pgp, liu2024_laformer, kim2021_lapred}, we also incorporate additional context information, such as static map information and the past trajectories of surrounding agents, into the model input.
A trajectory prediction model $f(x) = y $, parameterized by $\gW$, uses this input to estimate a future trajectory $y \in \R^{T_{out} \times F_{out}}$. 
Here, $T_{out}$ represents the prediction horizon, and $F_{out}$ is the number of output features to predict, such as coordinates.
Given the multi-modal nature of an agent's future behavior, an extended version of this model predicts multiple future trajectories. 
The distribution over potential future outcomes,  $p(y|x, \gW)$, can take various forms, such as a categorical distribution \cite{deo2022_pgp}, a mixture of Laplacians \cite{liu2024_laformer}, a \ac{GMM} \cite{nayakanti2023_wayformer}, or a non-parametric form \cite{jiang2023motiondiffuser}.
Finally, we define an ensemble \cite{zhou2012ensemble} as a set of $M$ trajectory prediction models. 
These models may have different parameterizations and could belong to different model families. 
The ensemble can be constructed using various techniques, such as Dropout \cite{gal16_dropout}, Stochastic Gradient Langevin Dynamics (SGLD) \cite{welling2011_sgld}, or deep ensembles \cite{Lakshminarayanan17_ensembles}. 
This ensemble introduces a distribution $q(\gW)$ over neural network parameters, which is an approximation to the true posterior $p(\gW | \mathcal{D})$ \cite{wilson2020bayesian}.

Our objective is to develop a method for uncertainty quantification to assess the trustworthiness of a model. However, the source of uncertainty is not always clear. 
On the one hand, high uncertainty may stem from novel, previously unseen traffic scenarios. 
On the other hand, randomness arising from unpredictable driver behavior can lead to multiple plausible predictions. 
While previous works such as \cite{gilles2022uncertainty} and \cite{janjovs2023bridging} do not distinguish between uncertainty types, we argue that decomposing uncertainty is crucial for understanding the sources of potential error in prediction, which in turn supports safer and more effective downstream decision making. Therefore, following concurrent literature \cite{der2009aleatory, huellermeier2021uqml}, we decompose uncertainty into epistemic and aleatoric components.

\subsection{Monte Carlo Approximation of the Conditional Entropy and Mutual Information as a Measure of Aleatoric and Epistemic Uncertainty}\label{sec:solution}

\begin{figure}
\centering
\includegraphics[width=1\linewidth]{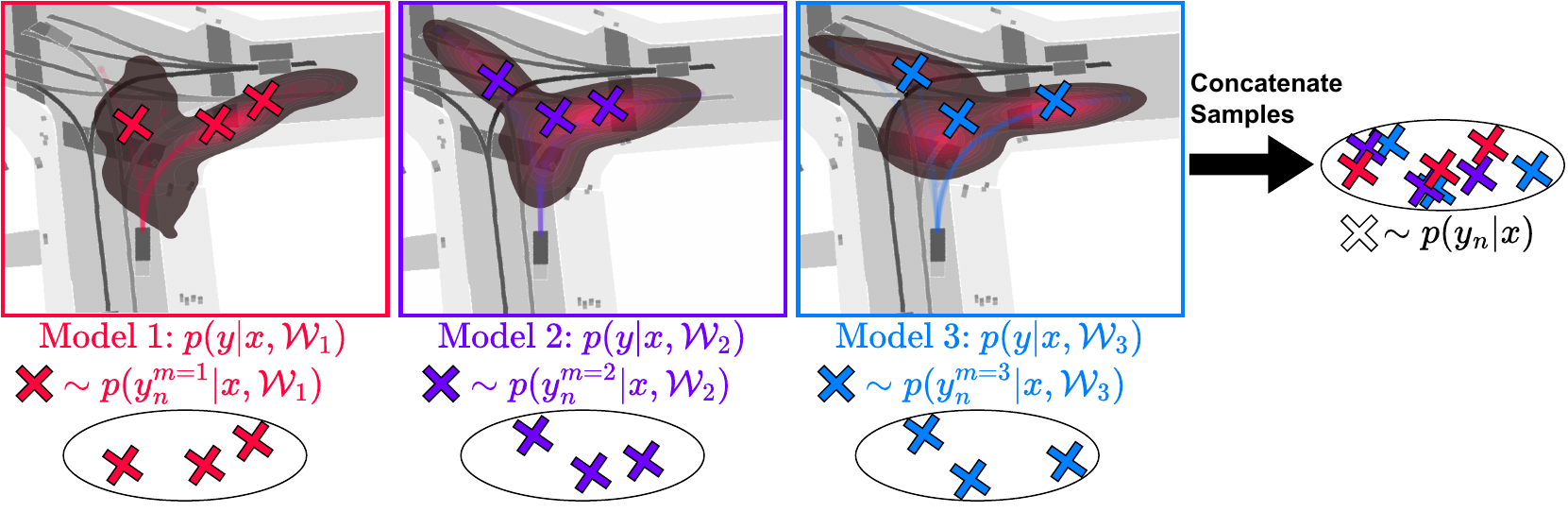}
\caption{\textbf{Generating samples for Monte Carlo approximation.} We fit a \ac{GMM} to the final positions of trajectories predicted by every member of our ensemble. Then, we sample from each \ac{GMM} to obtain per-model samples $y_n^m$ for calculating the term of aleatoric uncertainty. Finally, samples originating from all \ac{GMM}s are aggregated as $y_n$ for calculating the term of total uncertainty.}
\label{fig:method}
\vspace*{-.3cm}
\end{figure}

In quantifying uncertainty, we use entropy as a measure of total uncertainty. 
This allows us to frame our decomposition in terms of entropy components. 
Following \cite{Mobiny2019_dropconnect, depeweg2018decomposition}, we compute epistemic uncertainty as the difference between total and aleatoric uncertainty
\begin{equation}\label{eq:decomposition}
\underbrace{\rmI(y, \gW|x, \mathcal{D} )}_{\substack{\text{epistemic} \ \text{uncertainty}}} = \underbrace{\rmH(y | x,  \mathcal{D})}_{\substack{\text{total} \ \text{uncertainty}}} - \underbrace{\E_{p(\gW|\mathcal{D})}[\rmH(y | x, \gW)]}_{\substack{\text{aleatoric} \ \text{uncertainty}}}. 
\end{equation}
Above, \text{$\rmI(y, \gW|x,  \mathcal{D} )$} represents the mutual information between the model’s predictions and its parameters, while \text{$\rmH(y|x, \mathcal{D})$} denotes the total entropy of the predictive distribution.
The entropy of a distribution can be computed in closed form for simple cases, such as categorical distributions or univariate Gaussians. 
However, in trajectory prediction, the predictive distribution can take complex forms, such as a \ac{GMM} \cite{nayakanti2023_wayformer}, making closed-form solutions to Eq.~\ref{eq:decomposition} unavailable.
To address this, we use a Monte Carlo approximation. 
For a given input $x$, the entropy is approximated via  set of $N$ samples from the predictive distribution, $y_n \sim p(y|x, \mathcal{D})$ as 
\begin{align}
    \nonumber {\rmH}(y|x, \mathcal{D}) &= \E_{y}\left[ -\log p(y|x, \mathcal{D}) \right] \\
    &\nonumber \approx - \frac{1}{N}\sum_{n=1}^{N}{\log p(y_n|x, \mathcal{D})} \\
    &= \hat{\rmH}(Y|x, \mathcal{D}).
    \label{eq:monte_carlo}
\end{align}
Next, we replace the true posterior over neural network parameters  $p(\gW|\mathcal{D})$ with the approximate posterior $q(\gW)$. 
The approximate posterior is a discrete distribution over a set of $M$ neural network parameter values \(\gW_m\), allowing us to approximate the predictive distribution as
\begin{align}
\nonumber p(y|x, \mathcal{D})   &=\E_{p(\gW|\mathcal{D})}[p(y|x, \gW) ] \\
\nonumber &\approx \E_{q(\gW)}[p(y|x, \gW) ] \\
&= \frac{1}{M}\sum_{m=1}^M p(y|x, \gW_m).
\label{eq:predictive_discrete}
\end{align}
The choice of the model composition $q(\gW)$ significantly impacts the results, as different models may produce varied predictions, which will be explored further in Sec.~\ref{sec:results}. 
We then continue by inserting both Eq. \ref{eq:monte_carlo} and \ref{eq:predictive_discrete} into the original problem as defined in Eq. \ref{eq:decomposition}
\begin{align}
\nonumber \rmI(y,\!\gW| x,\! \mathcal{D})\!&\approx \hat{\rmH}(y|x, \mathcal{D}) - \E_{q(\gW)}[\hat{\rmH}(y|x,\gW)],\\ 
\nonumber &\stackrel{Eq.~\ref{eq:monte_carlo}}{=} -\frac{1}{N}\sum_{n=1}^{N}{\log p(y_n|x, \mathcal{D})} \\
\nonumber & \hspace{1cm} - \E_{q(\gW)}\left[ - \frac{1}{N} \sum_{n=1}^{N}{\log p(y_n|x,\gW)} \right] \\
\nonumber &\stackrel{Eq.~\ref{eq:predictive_discrete}}{=}  - \frac{1}{N} \sum_{n=1}^{N}{\log\!\left( \frac{1}{M}\! \sum_{m=1}^{M}{p(y_n|x,\gW_m)}\!\right)}\! \\
& \hspace{1cm}+\!  \frac{1}{M} \sum_{m=1}^{M}{\frac{1}{N}\!\sum_{n=1}^{N}{\log p(y_n^m|x,\gW_m)}}.
\label{eq:final}
\end{align}
Above, $y_n^m$ represents the $n$-th sample from the $m$-th model, i.e., $y_n^m \sim p(y|x, \gW_m)$. 
In contrast, $y_n$   represents the $n$-th sample from the predictive distribution after integrating out the weights, i.e., $y_n \sim p(y|x, \mathcal{D})$.
We visualize the sampling of $y_n$ in Fig. \ref{fig:method}. 
In essence,  we first collect equally-sized sets of $N'$ samples from each distribution $p(y|x,\gW_m)$, such that $N=N'\cdot M$. 
Concatenating them generates $N$ samples from the distribution \text{$p(y|x, \mathcal{D})$}, as the weights $\gW_m$ are equally weighted.

Our proposed approach formalized in Eq.~\ref{eq:decomposition}-~\ref{eq:final} assumes a generic form of the distribution $p(y|x,\gW_m)$. In practice, we use a continuous \ac{GMM} that is ubiquitous in trajectory prediction for \ac{AD}, see Sec.~\ref{subsec:related_work_motion_prediction}. 
Thus, we fit samples from a trajectory prediction model to a \ac{GMM}, or directly use the GMM if the predictor provides one. 
In Fig.~\ref{fig:method}, we visualize \ac{GMM}s fitted to the predictions from $M{=}3$ ensemble components, as well as samples from each \ac{GMM} over a two-dimensional grid.

\subsection{Discussion}\label{sec:limitations} The proposed approach effectively quantifies uncertainty in trajectory prediction. However, it is important to acknowledge several current limitations and potential solutions. One notable challenge is the burden of increased memory and computation, which may be prohibitive for real-time applications such as trajectory prediction. A potential solution to this limitation is ensemble distillation, which combines an ensemble of models into a single, more efficient model, significantly reducing computational overhead while maintaining comparable accuracy \cite{malinin2019ensemble}. A distillation approach for motion prediction models has been proposed in~\cite{ettinger2024scaling}. Alternatively, ensembles can be constructed by modifying only the final layer \cite{harrison2024variational}, further mitigating computational costs. These techniques offer a promising direction for future work, ensuring that the approach remains both efficient and performant.


\section{Experiments}
\label{sec:results}

In this paper, we introduce a novel information-theoretic approach to measure and decompose the uncertainty of the predictive distribution of trajectory prediction models in the \ac{AD} domain. 
We model the approximate posterior $q(\gW)$ over neural network weights via sampling-based methods, such as dropout \cite{gal16_dropout} and deep ensembles \cite{Lakshminarayanan17_ensembles}. 
For simplicity, we refer to any collection of neural networks as an ensemble.
Our experimental analysis is divided into four parts, where we explore both the uncertainty quantification capabilities of our method and the impact of different ensemble compositions.
First, in Sec.~\ref{subsec:exp1_corr}, we benchmark our method against an alternative approach to quantify the uncertainty on the original nuScenes dataset \cite{nuscenes}, which is a commonly used real-world trajectory prediction dataset for \ac{AD}. 
We measure the correlation between the uncertainty and the prediction error and explore how epistemic and aleatoric uncertainties complement each other.
In the subsequent parts, we create artificial \ac{OOD} scenarios by manipulating the nuScenes dataset in various ways. 
Specifically, we propose four different methods for manipulating the original nuScenes dataset
as described below:
\begin{itemize}
    \item RevertEGO: Revert the history of the target vehicle.
    \item ScrambleEGO: Randomly shuffle the history of the target vehicle.
    \item Blackout: Set 1/2 of the history to zero for the target and all surrounding vehicles.
    \item LaneDeletion: Randomly delete 3/4 of all lanes.
\end{itemize}
Beyond that, we consider combinations of manipulations.
In the second experimental part in Sec.~\ref{subsec:exp2_robustness}, we examine the robustness of various models and ensembles across different \ac{OOD} scenarios. 
We observe an overall increase in prediction error, indicating that our artificial \ac{OOD} scenarios are more challenging than the original dataset. 
In the third part in Sec.~\ref{subsec:exp3_quant}, we investigate how the correlation between uncertainty and prediction error is affected in these \ac{OOD} scenarios. 
Lastly, in Sec.~\ref{subsec:exp4_det}, we study whether we can detect \ac{OOD} scenarios by analyzing the different types of uncertainty.

Throughout our experiments, we use our novel method to measure the total uncertainty and decompose it into aleatoric and epistemic components to understand their relative importance. 
We generate trajectory predictions from the ensemble using the approach described in \cite{distelzweig2024trajectories}, which involves \ac{MBRM} to draw trajectories from an ensemble of prediction models. 
For single models, we generate trajectories via Topk sampling, which selects the most likely trajectories \cite{liu2024_laformer}.
We rely on LAformer~\cite{liu2024_laformer}, PGP~\cite{deo2022_pgp}, and LaPred~\cite{kim2021_lapred} to construct different ensembles of trajectory prediction models. 
These three models are among the best-performing models with available open-source implementations.
In our experiments, we evaluate different ensemble configurations, including deep ensembles, dropout ensembles, and single models. 
We use an ensemble size of three in all experiments; for deep ensembles, we sample three different models, and for dropout ensembles, we sample three different masks.
Prediction performance is assessed in terms of \acf{minADE} over $5$ proposals. 
The $\text{minADE}_5$ measures the average point-wise L2 distances between the predicted trajectories and the ground truth, returning the minimum over the  proposals~\cite{nuscenes}. 

\subsection{Correlation between Prediction Error and Different Uncertainty Types}
\label{subsec:exp1_corr}
Determining whether predictions can be trusted is crucial for deciding when to rely on the system or when the driver should take control. In this experiment, we analyze the correlation between different types of uncertainty and prediction error using the original nuScenes dataset.
More concretely, we compute the Pearson correlation coefficient $\rho$ between each type of uncertainty and the $\text{minADE}_5$. 
We benchmark our proposed method against~\cite{filos2020can}, which is an uncertainty quantification approach for planning. 
To the best of our knowledge, it is the only other architecture-agnostic method that addresses uncertainty quantification in the domain of autonomous driving.
More concretely, \cite{filos2020can} estimates uncertainty by computing the variance of the log-likelihood of future trajectories with respect to the parameters, i.e.,
 $\text{Var}_{q(\gW)}[\log p(y|x, \gW)]$. 
We report the $\text{minADE}_5$ values along with the correlation coefficient between different uncertainty types and the $\text{minADE}_5$ in Tab.~\ref{tab:benchmark_original_dataset}.

\begin{table*}
\caption{Correlation between Different Uncertainty Types and Prediction Error on the nuScenes Dataset.}
\vspace*{-0.2cm}
\label{tab:benchmark_original_dataset}
\centering
\begin{threeparttable}
\begin{tabular}{ll|cccc|ccc|ccc}
\toprule
&& \multicolumn{4}{c|}{Deep Ensembles} & \multicolumn{3}{c|}{Dropout Ensembles} & \multicolumn{3}{c}{Single Models} \\
\cmidrule(lr){3-12}
&& $1 \times$ & $3 \times$ & $3 \times$ & $3 \times$ & $3 \times$  &$3 \times$  & $3 \times$ & $1 \times$ & $1 \times$ & $1 \times$  \\
&& $ {\text{LP, LF, PGP}}$ & $ \text{PGP}$ & $ \text{LF}$ & $ \text{LP}$ &  PGP &  LF &  LP & PGP & LF & LP  \\
\midrule
\multicolumn{2}{c|}{$\text{MinADE}_5$} & 1.22 & 1.22 & 1.20 & 1.34 & 1.26 & 1.28 & 1.39 & 1.28 & 1.51 & 1.53 \\
\midrule
\parbox[t]{2mm}{\multirow{3}{*}{\rotatebox[origin=c]{90}{Ours}}}
&$\rho_{total}$ & 0.38 & 0.35 & 0.39 & 0.27 & 0.31 & 0.37 & 0.21 & 0.27 & 0.26 & 0.15 \\
&$\rho_{aleatoric}$ & 0.36 & 0.34 & 0.38 & 0.19 & 0.31 & 0.36 & 0.15 & 0.27 & 0.26 & 0.15 \\
&$\rho_{epistemic}$ & 0.28 & 0.23 & 0.25 & 0.28 & 0.21 & 0.28 & 0.23 & - & - & - \\
\midrule
RIP & $\rho_{epistemic}$  & 0.06 & 0.14 & 0.10 & 0.11 & 0.04 & 0.17 & 0.17 & - & - & - \\
\bottomrule
\end{tabular}
$\text{minADE}_5$ and Pearson correlation (higher is better) between $\text{minADE}_5$ and different uncertainty types on the original nuScenes dataset.  
We use sampling via \ac{MBRM} \cite{distelzweig2024trajectories} for ensembles and Topk for single models. LP~=~LaPred \cite{kim2021_lapred}, LF = LAformer \cite{liu2024_laformer}, PGP \cite{deo2022_pgp}, Dropout \cite{gal16_dropout}, RIP = Robust
Imitative Planning \cite{filos2020can}.
\end{threeparttable}
\vspace*{-.3cm}
\end{table*}

We first compare the correlation between the $\text{minADE}_5$ and different uncertainty types estimated by our method.
We observe that for all ensembles except $3 \times \text{LP}$, the total uncertainty has an equal or higher correlation with the prediction error than its individual components, i.e. the aleatoric and epistemic uncertainty. 
This suggests that both uncertainty sources are complementary.
When comparing ensembles against single models, we observe that all ensembles outperform the single models, as these models do not account for epistemic uncertainty.
Moreover, when comparing deep ensembles against dropout ensembles, we observe that the former offers a higher correlation coefficient. This indicates that deep ensembles quantify uncertainty more accurately than dropout, which is in line with the literature on uncertainty quantification with deep ensembles~\cite{Lakshminarayanan17_ensembles, durasov2021masksembles}. 
Lastly, we compare our method against the uncertainty quantification method proposed in~\cite{filos2020can}, i.e., \ac{RIP}. 
We observe that our uncertainty quantification method outperforms this approach for all model configurations. 
This is likely because \ac{RIP} is based on a heuristic, whereas our method takes a more comprehensive approach.
Overall, we observe that the uncertainty estimates obtained by our method provide a useful indication of whether we can trust our model's predictions or not.

\subsection{Robustness of Predictions in OOD Scenarios}
\label{subsec:exp2_robustness}
\begin{figure}
\centering
\includegraphics[width=1\linewidth]{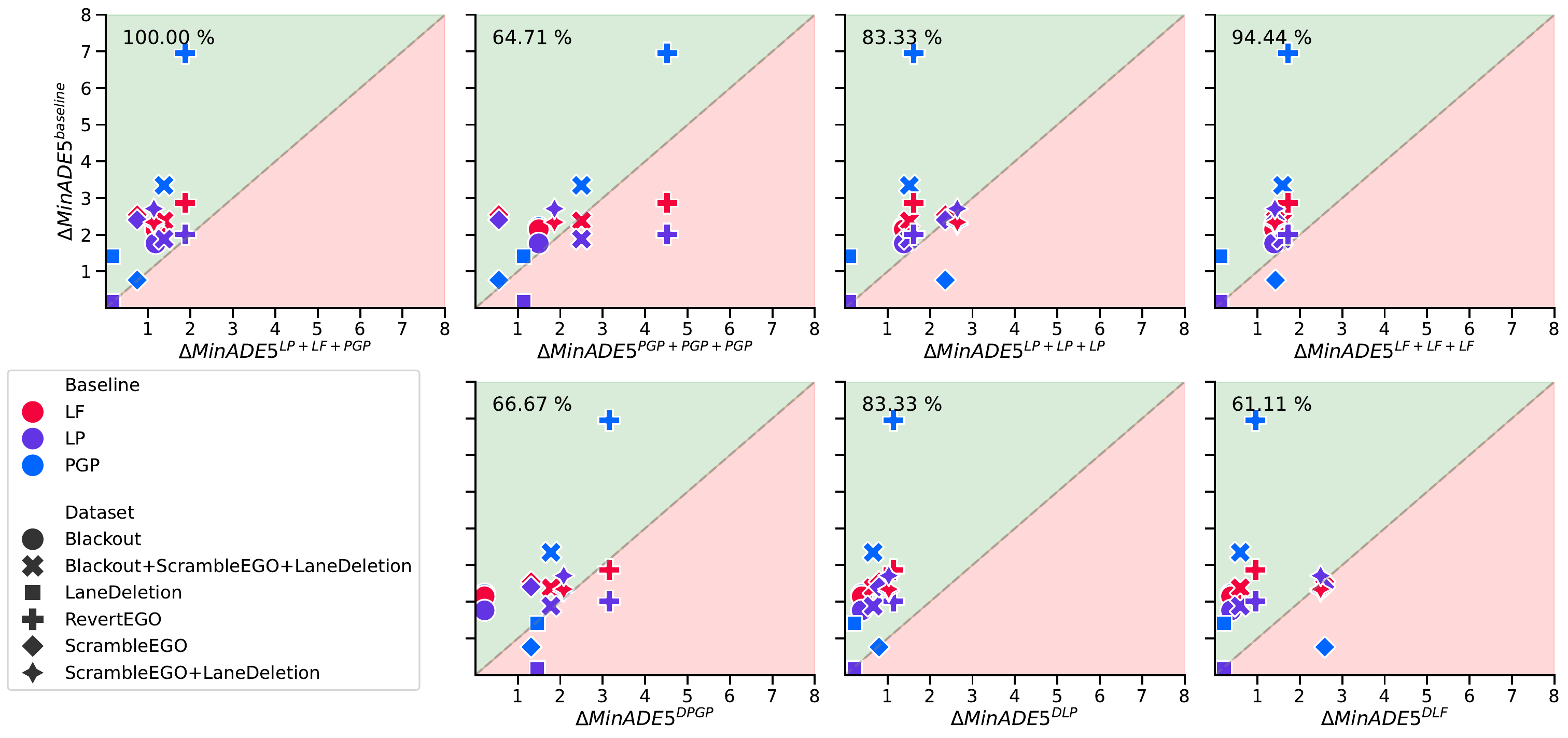}
\caption{Differences ($\Delta$) in $\text{MinADE}_5$ between the original dataset and the corresponding out-of-distribution dataset for baseline models (y-axis) and ensembles (x-axis). Different colors correspond to various baseline models, while different markers denote distinct dataset augmentations. Markers positioned in the red area (lower triangle) of each plot indicate that the ensemble exhibits a larger $\Delta \text{MinADE}_5$ compared to the baseline. Conversely, markers in the green area signify a smaller $\Delta \text{MinADE}_5$ for the ensemble. Percentages indicate how often the ensemble outperforms the baseline. Upper row represents deep ensembles and lower row Dropout ensembles.}
\label{fig:error_change_ood}
\vspace*{-.3cm}
\end{figure}

In the previous experiment, we analyzed the correlation between uncertainty and prediction error in \ac{ID} scenarios. 
We now shift our focus to examining whether prediction performance degrades in \ac{OOD} scenarios and to what extent. 
We report the changes in the $\text{minADE}_5$ metric with respect to the original dataset in Fig. \ref{fig:error_change_ood}. 

Overall, we observe that prediction error increases across all datasets in \ac{OOD} scenarios, indicating that our dataset augmentations create a more challenging evaluation setting. However, model ensembles consistently outperform individual models, as more than 50\% of the data points fall within the upper green triangle in Fig.~\ref{fig:error_change_ood} across all model configurations. This suggests that ensembles provide greater robustness and resilience in \ac{OOD} scenarios.
When comparing deep ensembles composed of the same model to their dropout-based counterparts, performance remains similar in terms of the fraction of data points in the green triangle. For example, the dropout ensemble outperforms deep ensembles for PGP, while LaPred exhibits equal performance across both configurations. In contrast, LAformer benefits more from deep ensembles.
Notably, when evaluating the mixed deep ensemble, which combines different models, we observe a substantial performance improvement, with all data points falling within the green triangle. 

\subsection{Quantifying the Uncertainty in OOD Scenarios}
\label{subsec:exp3_quant}
\begin{figure}
\centering
\includegraphics[width=1\linewidth]{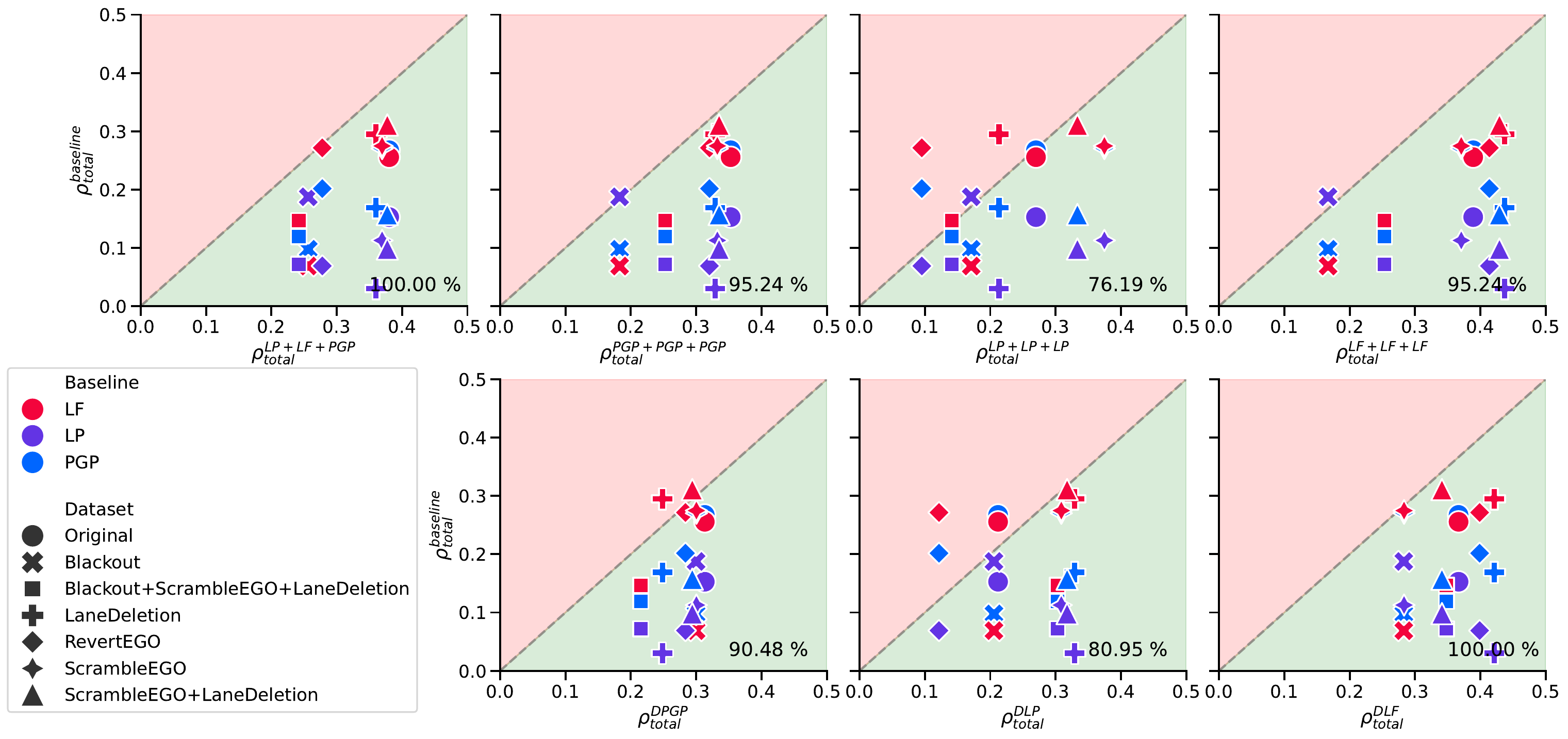}
\caption{Pearson correlation coefficient $\rho$ between total uncertainty and $\text{MinADE}_5$ for baseline models ($y$-axis) and ensembles ($x$-axis) over the validation set. Different colors represent various baseline models, while different markers indicate distinct dataset augmentations. Markers located in the red area (upper triangle) of each plot signify that the ensemble shows a lower correlation $\rho_{total}$ compared to the baseline. Conversely, markers in the green area (lower triangle) indicate a higher correlation for the ensemble. The numerical value in the bottom right corner of each plot represents the fraction of data points that fall within the green area.  Upper row represents deep ensembles and lower row Dropout ensembles.}
\label{fig:uncertainty_correlation_ood}
\end{figure}

In Sec. \ref{subsec:exp1_corr}, we investigated whether the uncertainty estimates from our method offer indications of the reliability of our model's predictions. 
However, it remains unclear if these findings are also applicable to \ac{OOD} scenarios. 
Specifically, can we trust our uncertainty estimates when encountering out-of-distribution inputs?
In this experiment, we analyze the correlation between uncertainty and prediction error in \ac{OOD} scenarios across different ensembles, and we compare these correlation coefficients with those obtained from the original dataset. 
We report the correlation coefficient between the total uncertainty and the $\text{minADE}_5$ in Fig. \ref{fig:uncertainty_correlation_ood}.

We first compare the correlation values from the original dataset represented by the circle marker in Fig.~\ref{fig:uncertainty_correlation_ood} with those from the OOD datasets represented by all other markers. 
The results present a mixed picture -- in some OOD scenarios, the correlation coefficient decreases while in others, it increases. 
Nevertheless, there is a general trend toward a decrease in the correlation coefficient in most OOD cases.

Interesting insights emerge when comparing the results of our approach with the results of the RIP uncertainty quantification approach~\cite{filos2020can}. 
Since RIP estimates only epistemic uncertainty, we evaluate the Pearson correlation coefficient between epistemic uncertainty and $\text{minADE}_5$ across all \ac{OOD} scenarios. Due to space constraints, we provide only a summary of the results.
In 41 out of 42 examined ensemble configuration and \ac{OOD} scenario combinations, our approach yields a higher correlation coefficient than RIP.
Notably, the average correlation increase is most pronounced in the mixed ensemble and LaPred ensemble, with improvements of 406\% and 434\% over RIP, respectively.
The smallest average increase is observed in the dropout PGP ensemble configuration, at 61\%. These results suggest that our method provides more robust uncertainty quantification, even in challenging \ac{OOD} conditions.


Next, we investigate whether using an ensemble of models is more beneficial than relying on a single model in \ac{OOD} scenarios. To evaluate this, we compare the correlation between uncertainty and prediction error for ensembles versus individual models in Fig.~\ref{fig:uncertainty_correlation_ood}. Our results consistently show that ensemble configurations outperform single-model baselines. This conclusion is reinforced by the fact that in every configuration, more than 50\% of the data points lie within the green triangle, indicating that ensembles provide a more reliable measure of uncertainty in \ac{OOD} scenarios compared to individual models.

We then compare three different ensemble configurations in Fig.~\ref{fig:uncertainty_correlation_ood}, specifically (i) dropout-based ensembles, (ii) deep ensembles composed of the same model, and (iii) mixed deep ensembles composed of Laformer, PGP, and LaPred. In two out of three cases, (i) outperforms (ii) in terms of the number of markers within the green triangle. However, when considering (iii), we observe that its markers are fully in the green triangle. This is a notable performance improvement compared to (ii) as well as (i), which manages to match the mixed ensembles only in a single configuration. These findings suggest that mixed ensembles, which benefit from increased model diversity, provide superior uncertainty quantification compared to other methods. This conclusion aligns with our previous results in Fig.~\ref{fig:error_change_ood}, where mixed ensembles consistently performed the best or matched other ensemble configurations in terms of robustness in \ac{OOD} scenarios. Therefore, we conclude that mixed deep ensembles are the most effective choice for handling \ac{OOD} scenarios.

\subsection{Detecting OOD Scenarios}
\label{subsec:exp4_det}
\begin{figure}
\centering
\includegraphics[width=1\linewidth]{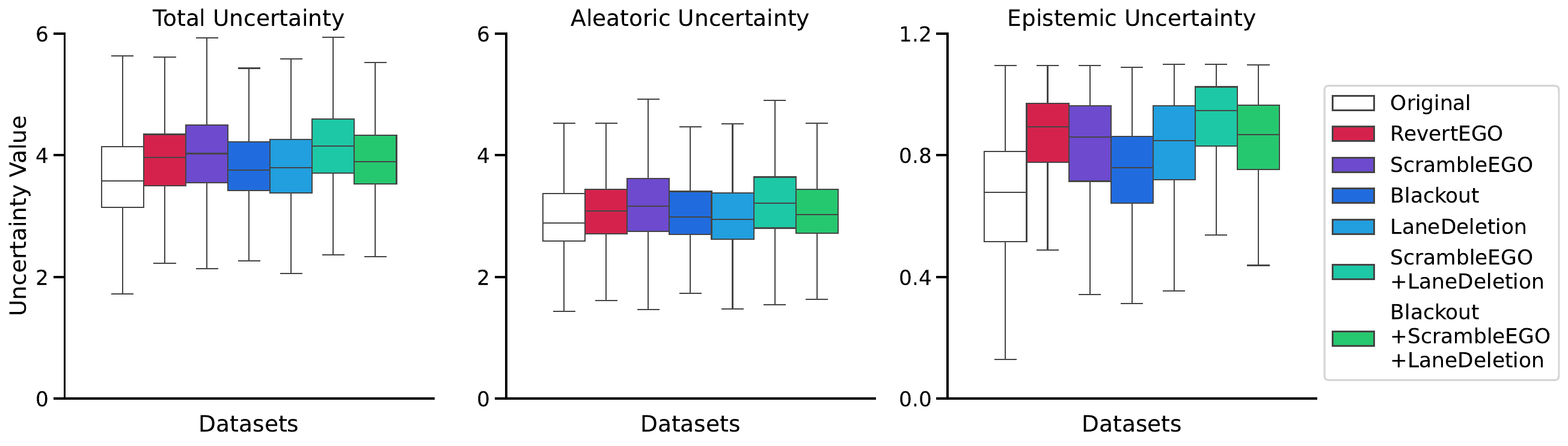}
\caption{Total, aleatoric, and epistemic uncertainties for a mixed ensemble ($1{\times} \text{LP, LF, PGP} $) for the original dataset as well as all out-of-distribution datasets.
}
\label{fig:avg_uncertainty_ood}
\vspace*{-.3cm}
\end{figure}

In this experiment, our objective is to determine whether \ac{OOD} scenarios can be reliably identified. Detecting such scenarios is crucial for safety, as it enables the autonomous system to alert the driver when intervention is necessary. Additionally, recognizing \ac{OOD} cases enhances the performance and robustness of an \ac{AD} system over time by facilitating the collection of challenging instances for re-training and evaluation.
We present the uncertainty values for different types of uncertainty in Fig. \ref{fig:avg_uncertainty_ood} for both the original nuScenes dataset and various \ac{OOD} scenarios.
For this analysis, we restrict our focus to a mixed deep ensemble consisting of LAformer, PGP, and LaPred, as this ensemble was favorable in terms of calibrations and robustness in previous experiments.

When analyzing epistemic uncertainty, we observe that \ac{OOD} scenarios exhibit a higher median value than the upper quartile of the original dataset, with the exception of the blackout scenario, where only the median of the original dataset is exceeded. 
In terms of aleatoric uncertainty, the median for \ac{OOD} scenarios consistently exceeds the median observed in the original dataset.
The total uncertainty follows a similar pattern to aleatoric uncertainty but exhibits a more pronounced difference between \ac{OOD} and \ac{ID} cases. These trends indicate that \ac{OOD} scenarios can be identified with highest confidence by assessing epistemic uncertainty, a finding that aligns with existing research in uncertainty quantification \cite{huellermeier2021uqml}.


\section{Conclusion}
\label{sec:conclusion}
Understanding and addressing uncertainty in probabilistic motion prediction for \ac{AD} remains a key challenge. This paper addresses this gap by proposing a general approach to quantify and decompose uncertainty using an information-theoretic framework.
We demonstrate that our estimates of aleatoric and epistemic uncertainty provide meaningful indicators of prediction error, making them reliable for assessing prediction performance. 
Through an extensive evaluation, we examine both in-distribution and out-of-distribution scenarios under various posterior assumptions. Overall, our approach advances principled uncertainty modeling in motion prediction for AD.

A promising future direction is to incorporate our uncertainty quantification framework into an integrated \ac{AV} prediction and planning system~\cite{teng2023planning, hagedorn2024integration}. Although the integration of trajectory-based motion prediction with planning is an open research problem~\cite{hagedorn2024integration}, solving it can be facilitated with comprehensive uncertainty estimates in decision making.


\bibliographystyle{IEEEtran}
\bibliography{IEEEabrv,bibliography_full}

\end{document}